\pdfoutput=1
\documentclass[letterpaper]{article}

\usepackage[english]{babel}
\usepackage[utf8x]{inputenc}
\usepackage[T1]{fontenc}
\usepackage[letterpaper,top=3cm,bottom=2cm,left=3cm,right=3cm,marginparwidth=1.75cm]{geometry}

\usepackage{amsmath,amsfonts}
\usepackage{graphicx}
\usepackage{subfig}
\usepackage{multirow}

\newtheorem{theorem}{Theorem}

\title{Lightweight Neural Networks}
\author{Altaf H. Khan\\
\texttt{altaf@altafkhan.com}\\
336 Eden Canal Villas, Lahore, Pakistan}

\begin{document}
\maketitle

\begin{abstract}
Most of the weights in a Lightweight Neural Network have a value of zero, while the remaining ones are either \(+1\) or \(-1\). These universal approximators require approximately 1.1 bits/weight of storage, posses a quick forward pass and achieve classification accuracies similar to conventional continuous-weight networks. Their training regimen focuses on error reduction initially, but later emphasizes discretization of weights. They ignore insignificant inputs, remove unnecessary weights, and drop unneeded hidden neurons. We have successfully tested them on the MNIST, credit card fraud, and credit card defaults data sets using networks having 2 to 16 hidden layers and up to 4.4 million weights.
\end{abstract}

\section{Lightweight Neural Networks}

\begin{figure}[t]\small
  \centering
  \subfloat{\includegraphics[width=0.5\textwidth, trim=0 0 5cm 1cm, clip]{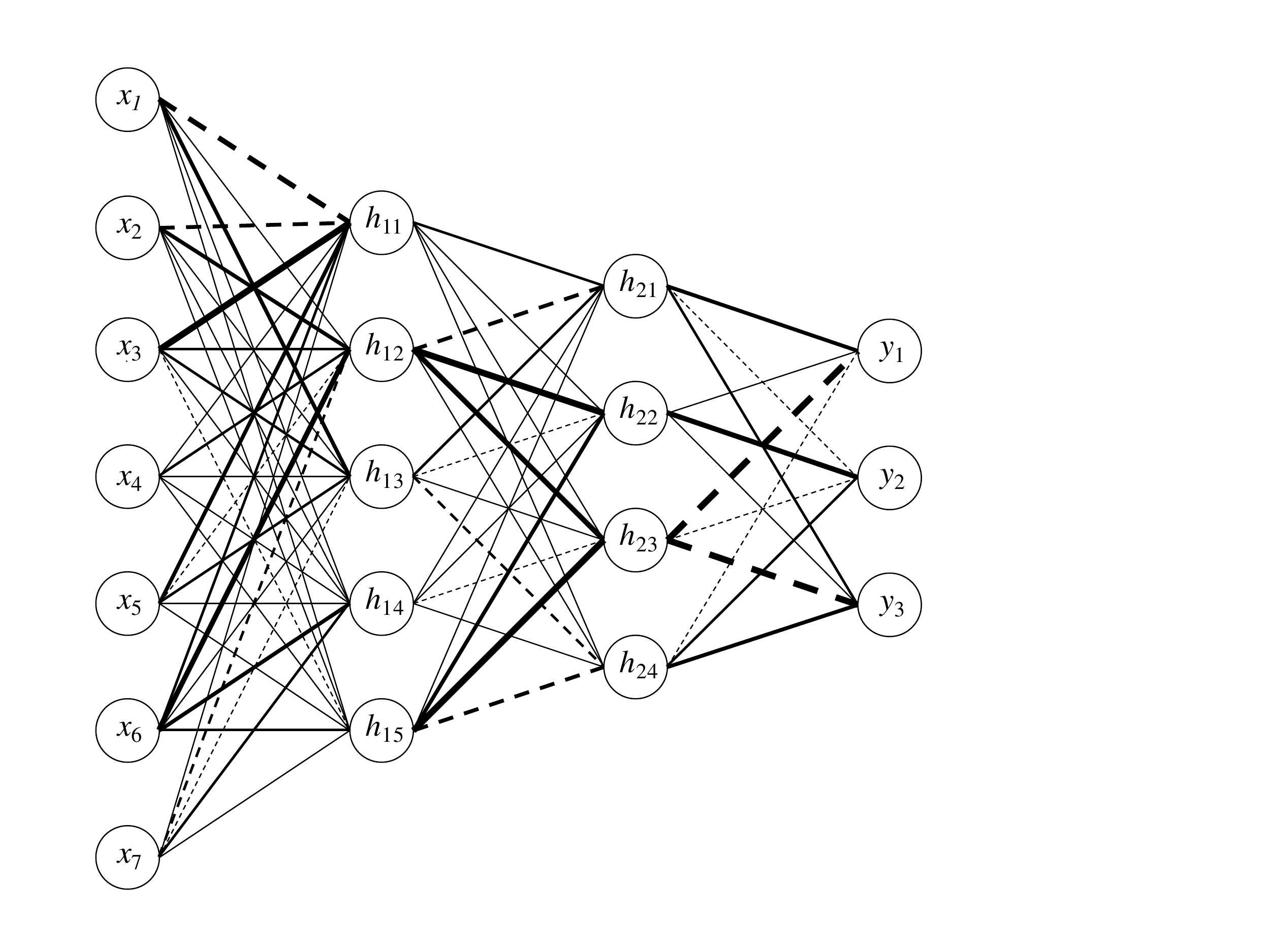}}
  \hfill
  \subfloat{\includegraphics[width=0.5\textwidth, trim=0 0 5cm 1cm, clip]{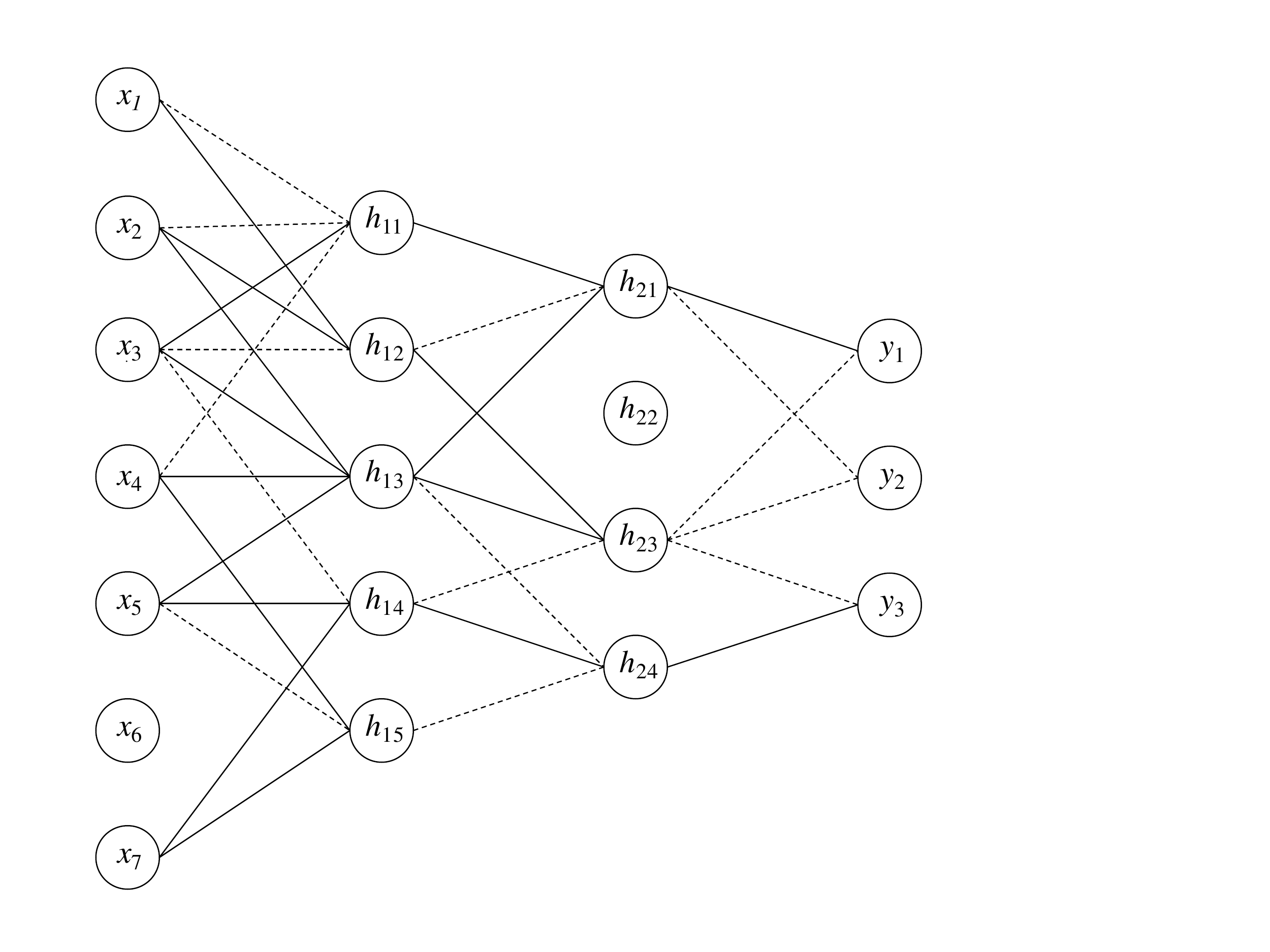}}
  \caption{\label{fig:cwn-lwn} Each CWN neuron is connected to all neurons in the preceding layer, whereas the LWN neurons have limited receptive fields. CWN weights have varying values (depicted by the thickness of the connections) and can be positive (solid lines) or negative (broken lines). The LWN weights have just two values, \(\pm1\).}
\end{figure}

Lightweight Neural Networks (LWN) are a subset of the conventional Continuous-Weight Networks (CWN). We call them \emph{lightweight} because the trained LWNs have weights that require approximately 1.1 bits/weight of storage and their forward-passes does not require floating-point multiplications. The key characteristic of LWNs is the sparsity of their weight matrices. Moreover, the non-zero weights of these matrices are limited to only two values: \(\pm1\) (see Figure~\ref{fig:cwn-lwn}). These networks were first introduced in 1996~\cite{khan1996feedforward, khan2002multiplier} as Multiplier-Free Networks and used training heuristics that were proposed in 1994~\cite{khan1994integer}. Due to the recent interest in similar networks~\cite{hwang2014fixed, li2016ternary, yin2017training, mellempudi2017ternary}, we present new results highlighting the sparsity of these networks, their natural inclination towards forming tight receptive fields, and their universal approximation capability.

\section{LWN and the Biological Neural Network}

Here we would like to highlight those aspects of the LWN neurons that make them more similar in structure and function to the biological neurons as compared with the CWN neurons. Consider an axon of a source biological neuron connecting to the dendrite of the target neuron at a synapse. One of two types of neurotransmitter chemicals (either for excitatory or inhibitory receptors) is released from the axon's side of the synapse whenever the source neuron is activated~\cite{bear2007neuroscience}. These chemicals then bind with receptors on the dendrite-side of the synapse, resulting in an increase (in case of an excitatory receptor) or decrease (inhibitory receptor) in the electrical potential on the membrane of the target neuron. The electrical potential of that membrane is the sum of contributions due to the firings of all neurons that are connected through synapses to the target neuron. When the membrane's electrical potential reaches a threshold value, the target neuron fires. The highlight of the above narrative is the absence of multiplication operations and the presence of only two synaptic values, excitatory (similar to the \(+1\) weight of LWN connections) and inhibitory(\(-1\) weight)\footnote{LWN neurons are different from the biological neuron in having bipolar outputs as well as bipolar inputs. In case of the negative-valued inputs, the roles of excitatory and inhibitory receptors are reversed.}.

\begin{figure}\small\centering
\caption{\label{fig:weights-to-neurons-ratio}Ratio of number of non-zero weights to number of functioning neurons, \(N_{\pm1}/M^{functioning}\), against number of functioning neurons in LWNs on the MNIST data set}

\includegraphics[width=0.75\textwidth]{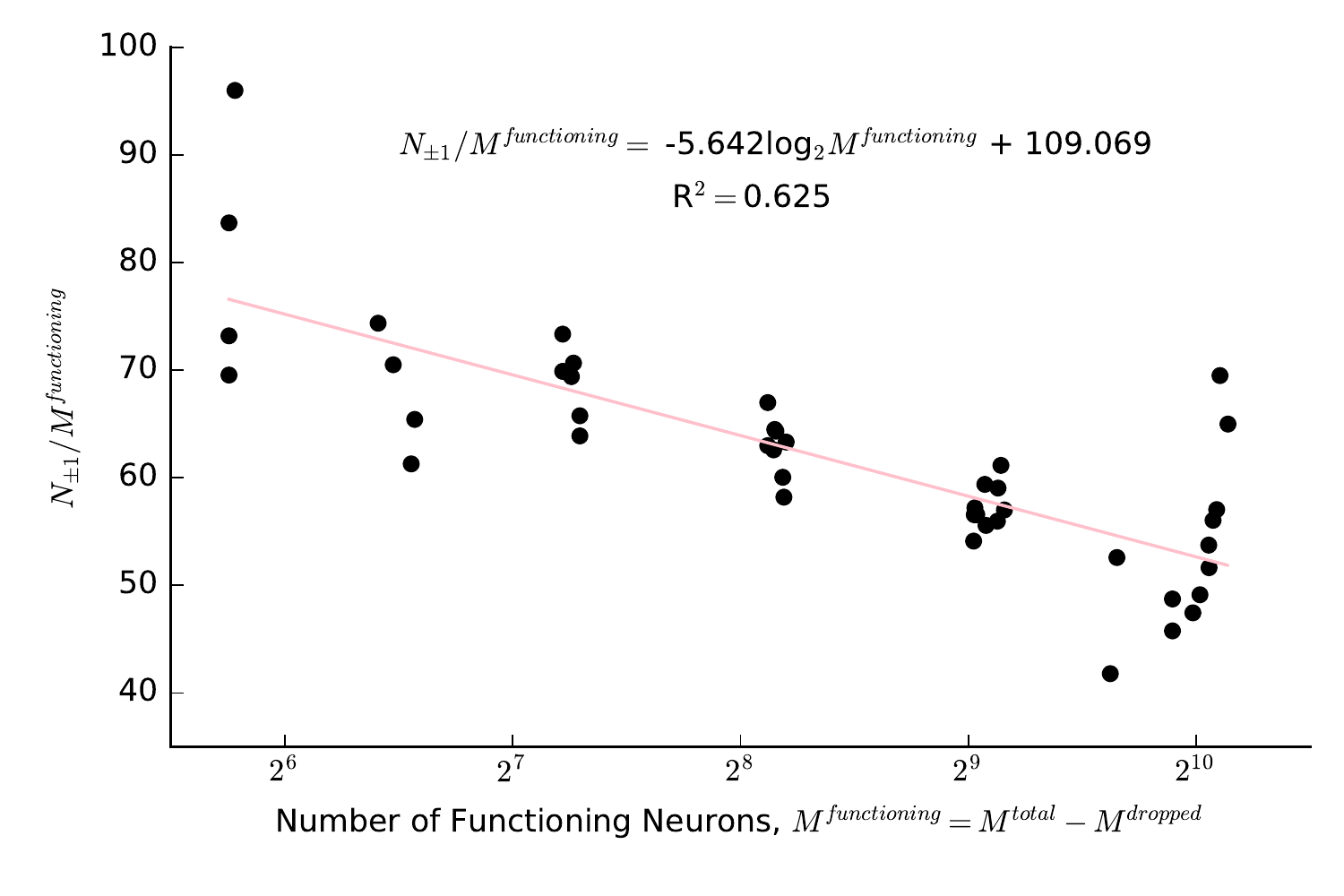}
\end{figure}

The connections-to-neurons ratio decreases with increase in the number of neurons in biological systems~\cite{ringo1991neuronal,cullen2010synapse}. LWNs exhibit the same characteristic (Figure~\ref{fig:weights-to-neurons-ratio}), but CWNs can not. The size of the receptive field, \(N_{RF}\), of a biological neuron is the fan-in of that neuron. Studies of processing in the visual cortex of animals show that \(N_{RF}\) varies among different types of neurons~\cite{hubel1968receptive}. The LWN is similar in structure in that each LWN neuron is a specialist (as compared to the generalist neurons of the CWN) as LWN neuron specializes in a particular subset of inputs. The \(N_{RF}\) for a conventional CWN is fixed for every neuron in every layer and is equal to the number of neurons in the preceding layer. For LWN, \(N_{RF}\) is much smaller and varies with the number of neurons in a layer.


LWN training is inspired by the synaptic pruning process in the biological brain~\cite{doi:10.1162/089976698300017124}: start with plenty; prune the excess off later. This natural phenomenon prunes, for example, 25-50\% synapses among humans as they approach adulthood, but does not reduce the number of neurons. For LWN, the training process prunes the initial count of the synapses (i.e.\ weights) by about 80-95\%, which, in many cases, results in the elimination of some of the neurons as well.

\section{Universal Approximation}

Although their weights are restricted to the set \(\{0, \pm 1\}\), LWNs' thresholds have no such constraints. Their activation functions are bounded and odd, and are confined to hyperbolic tangent in this discussion. With the help of these ingredients, we can create a one-dimensional bump of arbitrarily small height at an arbitrary location on the \(x\)-axis using expressions of the form
\[
\tanh(x - \rho) + \tanh(-(x - \rho) + \chi), 
\]
\noindent where \(\rho, \chi \in \mathbb{R}\). Summations of such bumps can be used to approximate arbitrary one-dimensional functions to any accuracy. We will now discuss the extension of this one-dimensional construction to denseness in \(C(\mathbb{R}^n)\). We start by a theorem due to Khan~\cite{khan2018weightless}:

\begin{theorem}[Weightless Neural Network Existence Theorem]
Summations of the form:
\[\sum_k^K \sigma\left( \sum_i^I x_i - c_k\right) + 
\sum_l^L \psi  \left( \sum_i^I x_i - c_l\right)\]
are dense in \(C(\mathbb{R}^n)\), where \(c_k, c_l \in \mathbb{R}\), \(\sigma(\cdot):\mathbb{R} \rightarrow \mathbb{R}\) is bounded and odd,  \(\psi(.) = \alpha \sigma(.)\), and \(\alpha \in \mathbb{R} \backslash \mathbb{Q}\).
\end{theorem}

\noindent The Weightless Neural Network is, on one hand, simpler than CWN in having unit-valued input- and output-layer weights, but on the other hand more complex in having two types of hidden neurons. The two types differing only by an irrational multiplication factor, \(\alpha\), in their activation functions.

Allowing the input- and output-layer weights to assume additional values of \(0\) and \(-1\) does not harm the density result, but may result in networks that train quicker and are more compact. The choice of multiplication factor \(\alpha\) is arbitrary, and can be chosen to be close to one. When simulated on a digital computer, that choice will become exactly one due to the limited precision of the computer~\cite{wray1995neural}, and the network will end up having only a single type of hidden neurons. This simplified configuration having \(I\) inputs and \(H\) hidden neurons can be written as:

\[\sum_k^H a_h \sigma \left(\sum_i^I b_{ih}x_i - c_h\right)\]

\noindent where \(a_h, b_{hi} \in \{0, \pm 1\}, \; c_h \in \mathbb{R}\). This expression depicts a network having a solitary layer of hidden neurons. To form multi-hidden-layer networks, layers comprising neurons identical to those in the first hidden-layer were employed. Moreover, for our simulations, we used output neurons which were identical to the hidden neurons. 

\section{Training Procedure}

Several approaches have been proposed for training neural networks with discrete weights. Hwang and Sung~\cite{hwang2014fixed} take a trained CWN, discretize its weights to ternary \(\{0, \pm1\}\) values, and retrain using backpropagation. They also restricted all signals to a depth of three bits. Mellempudi et al.\ \cite{mellempudi2017ternary} also start with a trained CWN and use a fine-grained ternarization method which exploits local correlations in the dynamic-range of the parameters to minimize the impact of discretization on the accuracy of the network. Li et al.\ \cite{li2016ternary} use the stochastic gradient descent method to train ternary-weight networks. They use ternary-weights during the forward and backward propagations, but not during the weight updates. Yin et al.\ \cite{yin2017training} gradually discretize the weights to zero or powers of two by minimizing the Euclidean distance between conventional weights and their closest discrete value during backpropagation.

We use the training heuristics proposed by Khan~\cite{khan1994integer} augmented by a new step with the aim of minimizing the error:

\[
E = E_o + E_w = \sum_{\textrm{all\,examples}} |\textbf{o}-\textbf{t}|^2 \; + \, \sum_{\textrm{all\,weights}} |w - \mathcal{Q}(w)|^2
\]

\noindent where \(\textbf{o}\) and \(\textbf{t}\) are the calculated and  desired output vectors, \(w\) the value of an individual weight, and \(\mathcal{Q}(\cdot)\) the weight discretization function. \(\mathcal{Q}(\cdot)\) is differentiable and the zeros of \((w - \mathcal{Q}(w))\) are \(\{0, \pm1\}\). The main point of the original heuristics was the sequential application of error-reduction and weight-discretization steps to the network in every training epoch. Both were based on steepest-descent, but weight-discretization was supplemented by an additional mechanism to take care of the weight-update paralysis. This paralysis was caused by the opposing weight-updates calculated by the error-reduction and weight-discretization steps. That additional mechanism - the black-hole mechanism - forced nearly-discrete weights to discrete values. The rate of weight-discretization and the radius of the black-hole grew as the error in the output of the network shrank.

\begin{figure}\small\centering
\caption{\label{fig:training-dynamics}Plot of training epochs against mean-squared error on training data, ratio of non-discrete weights to all weights, and test data miss-classifications. The bottom plot is the same as the top one, but with a magnified y-axis}
\includegraphics[width=0.7\textwidth]{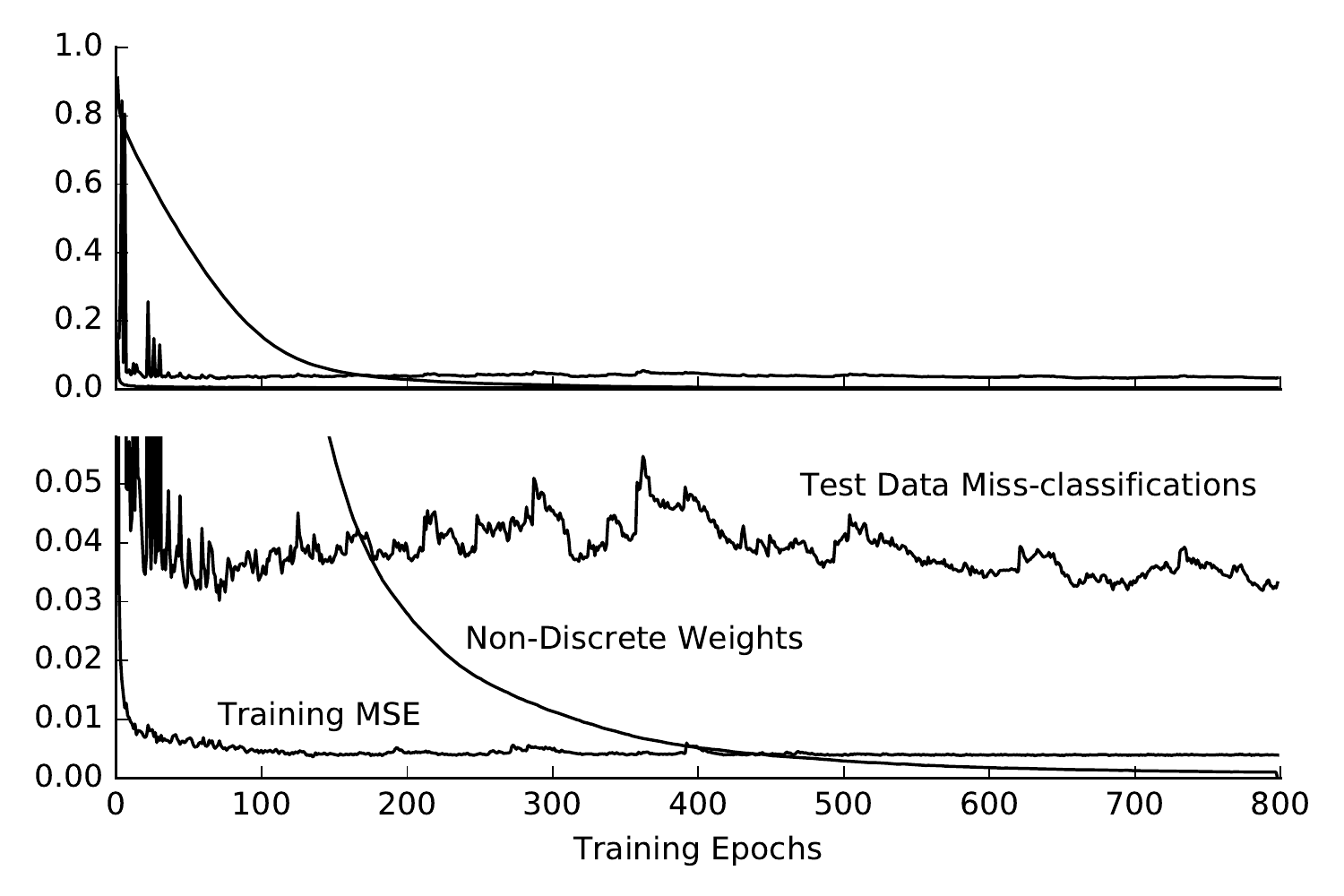}
\end{figure}

The black-hole mechanism worked well for shallow networks comprising hundreds of weights as discussed in~\cite{khan1994integer}, but failed to overcome the weight-update paralysis for deeper LWNs having thousands or millions of weights discussed in this paper. For such networks, we propose an additional mechanism that comes into play only when almost all weights are discrete. At that stage, all weights are rounded to their nearest discrete value from the set \(\{0, \pm1\}\). If that results in a network having acceptable test-data accuracy, training is concluded. Otherwise, the rounding step is rolled back and normal training is resumed with the pre-rounding weights. The resulting error plots for a typical LWN are shown in Figure~\ref{fig:training-dynamics}.

\section{MNIST Data Set Simulations}

MNIST is a well-known data set of images of handwritten digits~\cite{lecun2010mnist, lecun1990handwritten}. We used the version available through the TensorFlow machine learning library~\cite{abadi2016tensorflow}. That version includes 55,000 images in the training set and 10,000 in the test set. All images have been normalized, centered and transformed to a \(28\times28\) matrix of 32-bit floating-point numbers ranging from zero to one according to the gray-scale value of the associated pixel. The \(28\times28\) matrices have been flattened to 784-element vectors. Image labels are 10-element vectors with only a single element in a vector equal to one, and the rest having a value of zero. We used this data set as is and did not try to help the network configuration or the training process with any information about the original, 2-D nature of the image vector. This is because our interest is not to get the best possible result but to look at what the LWN learns with the simplest possible configuration. In this simplest configuration, all neurons in all layers are exactly the same and all layers are fully-connected to layer preceding them. The only factors that we varied were the number of hidden-layers and the number of neurons in each of those layers.

The summary of our results for two-hidden-layer LWNs is shown in Table~\ref{tab:mnist-results}. For each configuration, the results for the top-two LWNs with the best test-data accuracies are shown. The test-data accuracies range from 93.8 to 97.0\%. If we want to balance accuracy with size, we will choose the 784:256:128:10 LWN. It achieves an accuracy of 96.7\% with 0.23 million weights which compares favorably with 97.0\% for a much larger LWN having 1,152 extra neurons and 1.1 million additional weights. This result (Table~\ref{tab:compareCWN}) is similar to 96.95\% reported for a 784:300:100:10 CWN and 97.05\% reported for a 784:500:150:10 CWN~\cite{lecun1990handwritten}. We do not have much to say about the epochs required to achieve reasonable accuracy as that number is very much dependent on the training parameters. We did not make any effort to optimize the parameters for that purpose.

Figure~\ref{fig:training-dynamics} shows the mean-squared error on training data, the ratio of misclassifications to the total number of test data examples, and the ratio of non-discrete weights to the total number of weights as a function of training epochs. That figure clearly reflects the very deliberate slowness of the weight-discretization processes as compared with the error minimization process. Our intention was to find the continuous-weight error minimum first and then look for the discrete-weight minimum in its immediate vicinity. The lack of smoothness in the training error and test misclassification curves is due to the weight-discretization corrections. The smoothness of the non-discrete weight curve indicates the slowness of discretization.

\begin{table}\small\centering\setlength\tabcolsep{3.4pt} 

\caption{\label{tab:layer-wt-stats}\label{tab:mnist-results} Two-hidden-layer LWN results for the MNIST data}

\begin{tabular}{lrcrrcrrrc}\hline
\multirow{4}{*}{Configuration} &\multicolumn{1}{c}{\multirow{4}{*}{Epochs}} & Test &\multicolumn{1}{c}{\multirow{4}{*}{\(N\)}} &\multicolumn{1}{c}{\multirow{4}{*}{\(N_{±1}\)}} &&\multicolumn{3}{c}{Number of All-Zero Rows in}  & Maximum   \\
                               &&Data     &&&\multirow{2}{*}{\(N_0\)} &\multicolumn{3}{c}{the Weight Matrix of the}                                                                        &Fan-In   \\\cline{7-9}
                               &&Accuracy &&&                         &\multicolumn{1}{c}{\(1^\text{st}\) Hidden}  &\multicolumn{1}{c}{\(2^\text{nd}\) Hidden} &\multicolumn{1}{c}{Output} &of Output\\
                               &&(\%)     &&&(\%)                     &\multicolumn{1}{c}{Layer}                   &\multicolumn{1}{c}{Layer}                  &\multicolumn{1}{c}{Layer}  &Neurons  \\\hline
\multirow{2}{*}{784:30:14:10}    & 2,005  & 93.8   & \multirow{2}{*}{24,080}     & 3,756   & 84.4  & 124   & 0      & 0      & 3 \\
                                 & 3,670  & 93.8   &                             & 3,953   & 83.6  & 124   & 0      & 0      & 2 \\\hline
\multirow{2}{*}{784:32:16:10}    & 2,958  & 94.6   & \multirow{2}{*}{25,760}     & 4,520   & 82.5  & 111   & 0      & 4      & 4 \\
                                 & 6,625  & 94.8   &                             & 5,281   & 79.5  & 118   & 0      & 3      & 3 \\\hline
\multirow{2}{*}{784:64:16:10}    & 1,713  & 95.4   & \multirow{2}{*}{51,360}     & 6,322   & 87.7  & 66    & 0      & 5      & 2 \\
                                 & 1,129  & 95.1   &                             & 6,276   & 87.8  & 60    & 0      & 1      & 3 \\\hline
\multirow{2}{*}{784:64:32:10}    & 2,324  & 95.5   & \multirow{2}{*}{52,544}     & 6,216   & 88.2  & 65    & 1      & 10     & 2 \\
                                 & 615    & 95.2   &                             & 5,762   & 89.0  & 53    & 0      & 12     & 2 \\\hline
\multirow{2}{*}{784:128:16:10}   & 500    & 95.4   & \multirow{2}{*}{102,560}    & 10,620  & 89.6  & 27    & 0      & 1      & 3 \\
                                 & 407    & 95.3   &                             & 10,931  & 89.3  & 17    & 1      & 4      & 2 \\\hline
\multirow{2}{*}{784:128:32:10}   & 739    & 95.7   & \multirow{2}{*}{104,768}    & 10,415  & 90.1  & 20    & 0      & 21     & 2 \\
                                 & 1,356  & 95.6   &                             & 10,883  & 89.6  & 30    & 0      & 16     & 1 \\\hline
\multirow{2}{*}{784:128:64:10}   & 698    & 95.4   & \multirow{2}{*}{109,184}    & 10,032  & 90.8  & 19    & 0      & 45     & 2 \\
                                 & 472    & 95.6   &                             & 10,326  & 90.5  & 16    & 0      & 45     & 2 \\\hline
\multirow{2}{*}{784:256:16:10}   & 403    & 96.0   & \multirow{2}{*}{204,960}    & 18,625  & 90.9  & 1     & 5      & 2      & 2 \\
                                 & 683    & 95.9   &                             & 17,512  & 91.5  & 0     & 6      & 3      & 4 \\\hline
\multirow{2}{*}{784:256:32:10}   & 452    & 96.1   & \multirow{2}{*}{209,216}    & 17,715  & 91.5  & 4     & 0      & 15     & 1 \\
                                 & 672    & 96.1   &                             & 18,317  & 91.2  & 3     & 2      & 17     & 2 \\\hline
\multirow{2}{*}{784:256:64:10}   & 493    & 96.5   & \multirow{2}{*}{217,728}    & 18,336  & 91.6  & 3     & 0      & 44     & 3 \\
                                 & 840    & 96.4   &                             & 18,619  & 91.4  & 1     & 0      & 42     & 2 \\\hline
\multirow{2}{*}{784:256:128:10}  & 797    & 96.7   & \multirow{2}{*}{234,752}    & 16,996  & 92.8  & 2     & 1      & 109    & 2 \\
                                 & 488    & 96.3   &                             & 17,475  & 92.6  & 3     & 0      & 99     & 3 \\\hline
\multirow{2}{*}{784:512:16:10}   & 540    & 96.4   & \multirow{2}{*}{409,760}    & 29,867  & 92.7  & 0     & 13     & 3      & 2 \\
                                 & 432    & 96.2   &                             & 28,143  & 93.1  & 0     & 32     & 5      & 2 \\\hline
\multirow{2}{*}{784:512:32:10}   & 425    & 96.3   & \multirow{2}{*}{418,112}    & 29,474  & 93.0  & 0     & 16     & 17     & 2 \\
                                 & 547    & 96.1   &                             & 29,701  & 92.9  & 0     & 15     & 14     & 2 \\\hline
\multirow{2}{*}{784:512:64:10}   & 579    & 96.2   & \multirow{2}{*}{434,816}    & 30,012  & 93.1  & 0     & 2      & 48     & 2 \\
                                 & 549    & 96.0   &                             & 31,954  & 92.7  & 0     & 6      & 44     & 2 \\\hline
\multirow{2}{*}{784:512:128:10}  & 707    & 96.5   & \multirow{2}{*}{468,224}    & 31,288  & 93.3  & 0     & 0      & 80     & 2 \\
                                 & 856    & 96.3   &                             & 33,067  & 92.9  & 0     & 2      & 102    & 2 \\\hline
\multirow{2}{*}{784:512:256:10}  & 742    & 96.7   & \multirow{2}{*}{535,040}    & 32,556  & 93.9  & 0     & 0      & 207    & 2 \\
                                 & 783    & 96.1   &                             & 34,556  & 93.5  & 0     & 0      & 213    & 2 \\\hline
\multirow{2}{*}{784:1024:16:10}  & 390    & 96.2   & \multirow{2}{*}{819,360}    & 42,283  & 94.8  & 0     & 244    & 2      & 2 \\
                                 & 603    & 96.6   &                             & 32,934  & 96.0  & 0     & 258    & 4      & 2 \\\hline
\multirow{2}{*}{784:1024:32:10}  & 480    & 96.3   & \multirow{2}{*}{835,904}    & 46,399  & 94.4  & 0     & 96     & 18     & 2 \\
                                 & 619    & 96.2   &                             & 43,574  & 94.8  & 0     & 96     & 18     & 2 \\\hline
\multirow{2}{*}{784:1024:64:10}  & 507    & 96.6   & \multirow{2}{*}{868,992}    & 48,066  & 94.5  & 0     & 37     & 48     & 1 \\
                                 & 765    & 96.4   &                             & 50,851  & 94.1  & 0     & 15     & 48     & 2 \\\hline
\multirow{2}{*}{784:1024:128:10} & 799    & 96.1   & \multirow{2}{*}{935,168}    & 57,133  & 93.9  & 0     & 2      & 97     & 3 \\
                                 & 727    & 96.3   &                             & 54,955  & 94.1  & 0     & 2      & 96     & 2 \\\hline
\multirow{2}{*}{784:1024:256:10} & 1,129  & 96.3   & \multirow{2}{*}{1,067,520}  & 60,366  & 94.3  & 0     & 0      & 213    & 2 \\
                                 & 859    & 96.0   &                             & 62,125  & 94.2  & 0     & 0      & 201    & 3 \\\hline
\multirow{2}{*}{784:1024:512:10} & 1,520  & 96.2   & \multirow{2}{*}{1,332,224}  & 73,255  & 94.5  & 8     & 0      & 419    & 2 \\
                                 & 1,532  & 97.0   &                             & 76,459  & 94.3  & 10    & 0      & 446    & 2 \\\hline    
\end{tabular}
\end{table}

\begin{table}\small\centering
\caption{\label{tab:compareCWN} Best performing LWNs and CWNs on the MNIST data set. The boldface numbers highlight changes from the starting configurations.}

\begin{tabular}{lrrrrr}
\hline

\multirow{2}{*}{Network} &

\multirow{2}{*}
{\begin{tabular}[c]{@{}c@{}}Test Data\\Accuracy (\%)
\end{tabular}} &

\multirow{2}{*} 
{\begin{tabular}[c]{@{}c@{}}Configuration When\\Training Started \end{tabular}} &

\multicolumn{1}{c}{\multirow{2}{*}{\(N\)}} &

\multirow{2}{*} 
{\begin{tabular}[c]{@{}c@{}}Configuration at\\the End of Training \end{tabular}} &

\multicolumn{1}{c}{\multirow{2}{*}{\(N_{\pm1}\)}}
\\\\ \hline
CWN~\cite{lecun1998gradient} & 96.95 & 784:300:100:10 & 266,200 & n.a. & n.a.\\ 
CWN~\cite{lecun1998gradient} & 97.05 & 784:500:150:10 & 468,500 & n.a. & n.a.\\ 
LWN & 96.68 & 784:265:128:10 & 234,752 & \textbf{780}:\textbf{262}:\textbf{28}:10  & 16,996\\ 
LWN & 97.02 & 784:1024:512:10 & 1,332,224 & \textbf{774}:1024:\textbf{66}:10 & 76,459   \\ \hline
\end{tabular}

\end{table}
\subsection{Zero-Valued Weights}

\begin{figure}\small\centering 
\caption{\label{fig:H_1-H_2-Npm1} Relationship of the non-zero weights, \(N_{\pm1}\), and the number of neurons in the hidden layers, \( H_1\), \(H_2\), on the MNIST data set}
\includegraphics[width=0.75\textwidth, trim=0 0 0 2.5cm, clip]{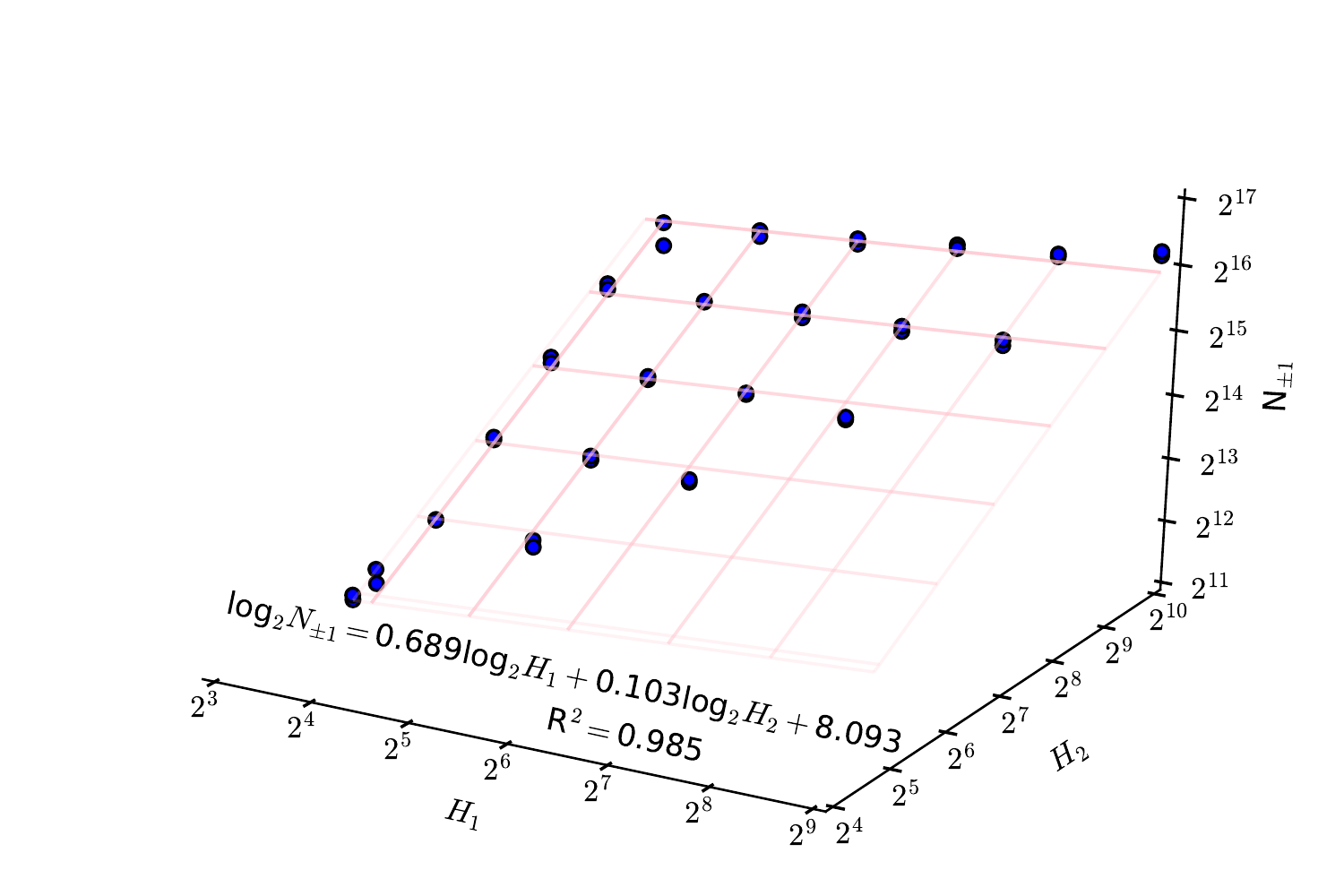}
\end{figure}

The sparsity, \(N_0\), of the two-hidden-layer LWNs ranged from 79.5 to 96\%, with the larger networks tending to have sparser weight matrices. The relationship of the dimensions of the hidden layers, \(H_1\) and \(H_2\), and the number of non-zero weights, \(N_{\pm1}\), is shown in Figure~\ref{fig:H_1-H_2-Npm1} on a log-log-log scale. \(N_{\pm1}\) seems to be mainly dependent on \( H_1\).

\subsubsection{Ignored Inputs}

The all-zero-rows column in Table~\ref{tab:layer-wt-stats} indicates that the LWN ignores some of the inputs when the number of neurons in the first hidden layer is inadequate. For the MNIST data set, the minimum layer size seems to be 256. This number is consistent with a minimum of 300 that was reported by LeCun et al.~\cite{lecun1998gradient}. For the 784:32:16:10 LWNs, the ignored pixels were in the top five and bottom three rows, and four left-most and three right-most columns. Whereas for the larger 784:128:64:10 LWNs the ignored pixels were in the top five and bottom row, and three left-most and two right-most columns. This indicates that the normalized MNIST images are slightly off-center and the LWNs are ignoring the nearly-white pixels around the digits due to their low information value.

\subsubsection{Receptive Fields}

Due to its fully-connected structure, the weights-to-neurons ratio for CWN increases with the number of neurons. The LWN restricts this rise in complexity by limiting each neuron to a narrow receptive field. Figure~\ref{fig:weights-to-neurons-ratio} shows the weights-to-neurons ratio as a function of the number of neurons in the network.

\begin{figure}\small\centering
\caption{\label{fig:H_1-N_RF} Relationship of the average size of the receptive fields, \(N_{RF}^{average}\), and the number of neurons in the first hidden-layer, \(H_1\), on the MNIST data set. The regression model excludes the first four points.}

\includegraphics[width=0.75\textwidth]{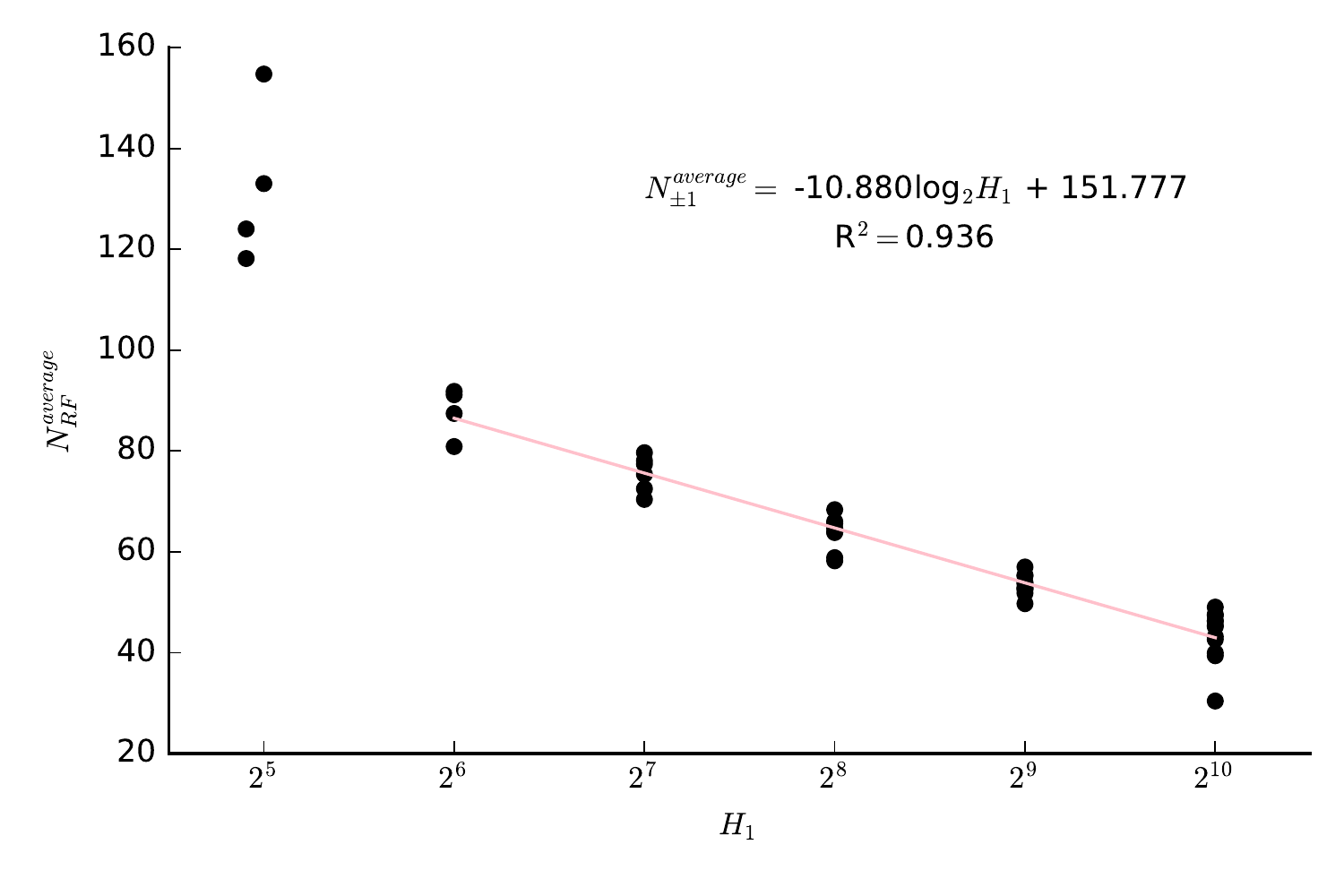}
\end{figure}

Variations in the average size of the receptive field of neurons in the first hidden-layer, \(N_{RF}^{average}\), as a function of the size of the hidden layer is shown in Figure~\ref{fig:H_1-N_RF}. Here, we excluded the smaller networks, those with 30-32 neurons in the hidden-layer, from the linear-regression model. The LWN reduces the burden of processing on each neuron by reducing their \(N_{RF}\) as more and more of them are added to the first hidden-layer.

We emphasize here that the LWN has no way of knowing that the MNIST data set comprises 2-D images. Neither does the LWN architecture include any custom-designed, small-receptive-field convolution layers of the Convolution Neural Network (CNN) \textit{\`a\ la} LeCun et al.~\cite{lecun1990handwritten}. It, however, tends to drop non-crucial inputs from the receptive-field of neurons by setting the corresponding weight to zero. This behavior is true for almost all neurons in all layers of the network. In contrast with the CNN, the size of the receptive field varies from neuron to neuron in an LWN and is not a parameter of the training process, but is arrived at naturally as a consequence of the training process.
Receptive-fields are clearly present in biological systems for the processing of visual, aural and touch and possibly other stimuli. The design of CNN, for example, is inspired by those systems.  There may be other data-processing situations where the receptive-fields may not be that obvious. The LWN can automatically discover and leverage them even in those situations.

\subsubsection{Dropped Hidden Neurons}

Some of the hidden neurons had all their weights set to zero as a consequence of the training regimen. The number of such dropped neurons, indicated by the number of all-zero rows in the weight matrices, is related to the mismatch between the number of neurons in consecutive layers. Table~\ref{tab:layer-wt-stats} makes this clear: greater the mismatch, the greater the number of dropped hidden neurons.

Tables~\ref{tab:layer-wt-stats} also shows that, in 70\% of the cases, a hidden-layer neuron is connected to at most two output neurons. This indicates the efficient distribution of the image recognition task among the hidden neurons. That is, the task is distributed in such a way that each neuron has a distinct and focused responsibility.  Here the size of the hidden-layer does not seem to be much of a factor. In the most extreme case, there are 446 dropped neurons in the second hidden layer. Of the remaining 66, 61 are connected only to single output neuron, while the other five are connected to two of the output neurons. The \(N_{RF}^{average}\) for the output neurons was 7.1.



\subsubsection{Weight Storage}

Each LWN weight requires at most two binary-bits for storage. However, using variable-length coding (e.\,g.\ Huffman coding~\cite{huffman1952method}), a weight matrix that is 90\% sparse will require only 1.1 bits/weight. This number approaches one as sparsity increases. Table~\ref{tab:storage} shows a comparison of storage requirements for various weight depths, but excludes the effect of thresholds on storage requirements. 32- and 8-bit thresholds for the 784:1024:512:10 configuration consume 48 and 12 kB, respectively. We have yet to try LWN training with 8-bit thresholds, but we expect it to work due to the results reported by Vanhoucke et al.~\cite{vanhoucke2011improving}.

\begin{table}\small\centering
\caption{\label{tab:storage} Weight storage comparison for the MNIST data set for the 784:1024:512:10 configuration that has \(N=1,332,224\) weights}
\begin{tabular}{rlrr}
\hline

\(N_{\pm1}/N\) & Storage Scheme & Bits/Weight & Required Storage (kB) \\ \hline
\rule{0pt}{1.2em}n.a. & Conventional & 32    & 5,204 \\ 
n.a.   & Conventional                & 8     & 1,301 \\ 
0.943  & Conventional                & 2     & 325 \\
0.943  & Huffman coding              & 1.057 & 172 \\ \hline

\end{tabular}
\end{table}

\subsubsection{Forward Pass}
LWNs posses an efficient forward pass because multiplications in all neurons are trivial in nature as weights are restricted to the set \(\{0, \pm 1\}\). Moreover, due to the high sparsity of weight matrices, even that trivial operation is not necessary most of the time. The thresholds are 32-bit floating-point numbers, but they are never part of a multiplication operation. Therefore, the LWN does not require the floating-point multiplication operation at all.

\subsection{Training Deeper Networks}

We trained several LWNs with 4-16 hidden-layers of varying sizes to see if our training heuristics suffered from the vanishing (and/or exploding) gradient problem~\cite{hochreiter1991untersuchungen, schmidhuber2015deep}. Results are shown in Table~\ref{tab:deeper-nets}. We did these training runs using the same training parameter values (e.g.\ learning rate, momentum, discretization rates) as those of the 2-hidden-layer LWNs. Our focus was not on obtaining the best performing LWNs, but to see if we were able to train theses deeper LWNs at all. Surprisingly, these training experiments did not present symptoms of the vanishing or exploding gradient. This is probably due to the restriction on the magnitude of the weights.

\begin{table}\small\centering\setlength\tabcolsep{4pt} 
\caption{\label{tab:deeper-nets} MNIST training results for LWNs having 4-16 hidden layers}
\begin{tabular}{lrrcrrc} \hline

\multirow{2}{*}{Configuration} &

\multirow{2}{*}
{\begin{tabular}[c]{@{}c@{}}Hidden\\Layers
\end{tabular}} &

\multirow{2}{*}{Epochs} &

\multirow{2}{*}
{\begin{tabular}[c]{@{}c@{}}Test Data\\Accuracy (\%)
\end{tabular}} &

\multicolumn{1}{c}{\multirow{2}{*}{\(N\)}} &

\multicolumn{1}{c}{\multirow{2}{*}{\(N_{\pm1}\)}} & 

\multirow{2}{*}{\begin{tabular}[c]{@{}c@{}}\(N_0\)\\(\%)
\end{tabular}} \\\\\hline

\multirow{2}{*}{784:1024:512:128:64:10} & \multirow{2}{*}{4} & 339 & 96.3 & \multirow{2}{*}{1,499,776} & 116,244 & 92.2 \\
 &  & 469 & 97.0 & & 98,018 & 93.5 \\\hline
\multirow{2}{*}{784:256:128:64:32:16:10} & \multirow{2}{*}{5} & 481 & 96.8 & \multirow{2}{*}{244,384} & 19,364 & 92.1 \\
 &  & 874 & 96.9 & & 19,506 & 92.0 \\\hline
\multirow{2}{*}{784:512:256:128:64:32:10} & \multirow{2}{*}{5} & 321 & 96.3 & \multirow{2}{*}{575,808} & 46,192 & 92.0 \\
 &  & 330 & 96.0 & & 48,010 & 91.7 \\\hline
 
\multirow{2}{*}{784:2048:1024:512:256:128:64:32:16:10} & \multirow{2}{*}{8} & 630 & 96.0 & \multirow{2}{*}{4,401,824} & 255,229 & 94.2 \\
 &  & 1,081 & 95.8 & & 264,572 & 94.0 \\\hline
 
\multirow{2}{*}{784:256:128:\{5\(\times\)64\}:32:16:10} & \multirow{2}{*}{9} & 771 & 96.5 & \multirow{2}{*}{260,768} & 21,664 & 91.7 \\
 &  & 655 & 96.6 & & 21,198 & 91.9 \\\hline
\multirow{2}{*}{784:256:128:\{12\(\times\)64\}:32:16:10} & \multirow{2}{*}{16} & 822 & 95.7 & \multirow{2}{*}{289,440} & 25,732 & 91.1 \\
 &  & 862 & 95.7 & & 26,280 & 90.9 \\\hline
\end{tabular}
\end{table}

\section{Credit Card Fraud and Defaults Simulations}

After extensive simulations on the MNIST data, we looked at credit card fraud~\cite{dal2015calibrating} and default~\cite{yeh2009comparisons} data to further validate the viability of the LWN. A key characteristic of these data sets is their class imbalance. This issue can be addressed by undersampling the majority class or oversampling the minority class. We tried oversampling in two different ways: simple repetition of the minority class examples and {\em Synthetic Minority Oversampling Technique} (SMOTE)~\cite{chawla2002smote}. For these simulations, we normalized the continuous features to zero-mean and unit-variance and clipped them to the range \([-1,1]\). All binary features were mapped to \(\{-1,1\}\). We then split the data into training (70\%) and testing (30\%) sets using stratified sampling. We assigned separate outputs to each of the two classes. The higher of the two outputs was considered the winning class while testing. This methodology avoids the work required to find the most suitable output threshold (in case of a single output neuron) at which the classification switches from one class to the other.

The credit card fraud data set contained 29 features and consisted of 284,807 transactions, out of which 0.17\%  were fraudulent. Our best performing network had an F1 score~\cite{fawcett2006introduction} of 0.83 with oversampled training data. Pozzolo et al.\ have reported G-mean scores of 0.944, 0.770, and 0.794 on this data set using Logit Boost, Random Forests, and Support Vector Machine, respectively~\cite{dal2015calibrating}. It is difficult to compare these results with our G-mean score of 0.900 due to, among other things, differences in data pre-processing, training/testing data splits, and classification thresholds. 

\begin{table}\small\centering
\setlength\tabcolsep{6pt} 
\centering
\caption{Test data results for the credit card fraud data.}
\label{tab:ccfraud}
\begin{tabular}{lcccccccc}\hline

\multirow{2}{*}{\begin{tabular}[c]{@{}l@{}} Sampling\\Method\end{tabular}}
            & \multirow{2}{*}{Configuration}
                                                  & \multirow{2}{*}{\(N\)}
                                                                             & \multirow{2}{*}{\(N_{\pm1}\)} 
                                                                                     & \multirow{2}{*}{\begin{tabular}[c]{@{}c@{}} \(N_0\)\\(\%) \end{tabular}}
                                                                                            & \multirow{2}{*}{\begin{tabular}[c]{@{}c@{}} Accuracy\\(\%)\end{tabular}}
                                                                                                   & \multirow{2}{*}{Precision} 
                                                                                                           & \multirow{2}{*}{Recall} 

                                                                                                                   & \multirow{2}{*}{\begin{tabular}[c]{@{}c@{}} F1\\Score\end{tabular}} 

\\\\\hline
Undersampled & \multirow{3}{*}{29:256:128:64:32:2} & \multirow{3}{*}{50,496} & 1,710 & 96.6 & 97.7 & 0.058 & 0.804 & 0.109 \\
Oversampled  &                                     &                         & 6,389 & 87.3 & 99.9 & 0.857 & 0.811 & 0.833 \\
SMOTE        &                                     &                         & 7,655 & 84.8 & 99.9 & 0.829 & 0.818 & 0.823 \\
\hline
\end{tabular}
\end{table}

The credit card defaults data set contained 23 features and consisted of 30,000 cases, out of which 22\% were defaults. We replaced categorical features with separate features for each category. The three sampling techniques, in this case, resulted in very similar F1 scores (Table~\ref{tab:ccdefault}) due to the relatively lower imbalance among the two classes. Yeh et al.\ have reported  83\% accuracy using the neural network classifier on this data set. It is not possible to compare this result with our result of 68.5\% due to lack of adequate information about their training and testing processes.

\begin{table}\small\centering
\setlength\tabcolsep{3.25pt} 
\centering
\caption{Test data results for the credit card default data.}
\label{tab:ccdefault}
\begin{tabular}{lcccccccc}\hline

\multirow{2}{*}{\begin{tabular}[c]{@{}l@{}} Sampling\\Method\end{tabular}}
            & \multirow{2}{*}{Configuration}
                                                  & \multirow{2}{*}{\(N\)}
                                                                             & \multirow{2}{*}{\(N_{\pm1}\)} 
                                                                                     & \multirow{2}{*}{\begin{tabular}[c]{@{}c@{}} \(N_0\)\\(\%) \end{tabular}}
                                                                                            & \multirow{2}{*}{\begin{tabular}[c]{@{}c@{}} Accuracy\\(\%)\end{tabular}}
                                                                                                   & \multirow{2}{*}{Precision} 
                                                                                                           & \multirow{2}{*}{Recall} 

                                                                                                                   & \multirow{2}{*}{\begin{tabular}[c]{@{}c@{}} F1\\Score\end{tabular}} 

\\\\\hline
Undersampled & \multirow{3}{*}{32:512:256:128:64:32:16:8:4:2} & \multirow{3}{*}{191,144} & 22,528 & 88.2 & 68.5 & 0.365 & 0.571 & 0.445 \\
Oversampled  &                                     &                         & 37,648 & 80.3 & 68.4 & 0.368 & 0.596 & 0.455 \\
SMOTE        &                                     &                         & 46,449 & 75.7 & 67.5 & 0.359 & 0.595 & 0.448 \\
\hline
\end{tabular}
\end{table}

\section{Conclusions}

LWNs do not have weights in the conventional sense, just excitatory and inhibitory connections of unit strength. They can approximate any continuous function with any accuracy. They require modest storage and have a multiplication-free forward pass, rendering them suitable for deployment on inexpensive hardware. Their sparse weight matrices loosen the coupling among the layers, making the LWN more tolerant to failure of individual neurons. In a CWN, the learned information is distributed over all weights. In an LWN, the picture is less fuzzy and the localized nature of computation is much more obvious due to the presence of a large number of zero-valued weights. For image processing, a CWN does not scale well with increases in the resolution of the images due to CWN's fully-connected structure. An LWN, on the other hand, scales much better due to the sparsity of its weight matrices. The small magnitude of LWN weights should result in smooth mappings~\cite{bishop1995neuralSmoothMapping} and the small number of non-zero weights should result in low generalization error~\cite{moody1992effective}.

The LWN learning process automatically drops insignificant inputs, unnecessary weights, and unneeded hidden-neurons. That process is relatively complex and slow, but results in networks that are almost as accurate as the CWNs but have much lower information complexity. It is conjectured that those accuracies can be matched by using bigger networks. This can be understood by considering the limited number of angles at which an LWN neuron can draw classification boundaries as opposed to the CWN neuron that can draw those boundaries at an arbitrary angle. A superposition of two or more LWN neurons can, however, approximate those arbitrary boundaries, but the added complexity may not justify the minuscule improvement in approximation accuracy.

At this time, the LWN neurons have unrestricted thresholds. We are exploring if the magnitude of these thresholds can be restricted to an arbitrary value~\cite{stinchcombe1990approximating}, e.g.\ 1, which may lead to more efficient storage. In its current form, the LWN training process applies the same training operation to every weight irrespective of the value of the weight. We are trying to see if the training process can be made more efficient by sometimes ignoring those weights that have not deviated from a zero value for several epochs.

\bibliographystyle{ieeetr}
\bibliography{nnet.bib}

\end{document}